\documentclass[10pt,journal,compsoc]{IEEEtran}

\usepackage{subeqnarray}

\usepackage{amsmath,graphicx}
\usepackage[latin1]{inputenc}
\usepackage{subeqnarray}
\usepackage{amsfonts}
\usepackage{amssymb}
\usepackage{graphicx}
\usepackage{color}
\makeatletter
\newcommand{\thickhline}{%
    \noalign {\ifnum 0=`}\fi \hrule height 1pt
    \futurelet \reserved@a \@xhline
}
\usepackage{graphicx,amssymb,amstext}
\usepackage{amsmath}
\usepackage{grffile}
\usepackage{xcolor}
\usepackage{hyperref}
\usepackage{setspace}
\usepackage{fancyhdr}
\usepackage{tabularx}
\usepackage{mathdots}
\usepackage{bm}
\usepackage{epstopdf}
\usepackage{makecell}
\usepackage{multirow}
\usepackage{caption}
\usepackage{array}
\usepackage{subcaption}
\usepackage{dblfloatfix}

\newif\ifarial
\hypersetup{colorlinks=true, linkcolor=blue!50!black}
\usepackage[linesnumbered,ruled]{algorithm2e}
\usepackage{algpseudocode}
\begin{document}
%
\title{Deep Multi-view Learning to Rank}
%
%
%
\author{Guanqun Cao, Alexandros Iosifidis, Moncef Gabbouj, 
Vijay Raghavan, Raju Gottumukkala}
\setlength{\belowcaptionskip}{-8pt}
\newcommand{\Tr}{\operatorname{Tr}}
\newcommand{\X}{\mathbf{X}}
\newcommand{\Y}{\mathbf{Y}}
\newcommand{\Z}{\mathbf{Z}}
\newcommand{\Sb}{\mathbf{S}}
\newcommand{\W}{\mathbf{W}}
\newcommand{\D}{\mathbf{D}}
\newcommand{\V}{\mathbf{V}}
\newcommand{\I}{\mathbf{I}}
\newcommand{\e}{\mathbf{e}}
\newcommand{\x}{\mathbf{x}}
\newcommand{\y}{\mathbf{y}}
\newcommand{\A}{\mathbf{A}}
\newcommand{\B}{\mathbf{B}}
\newcommand{\Pb}{\mathbf{P}}
\newcommand{\Q}{\mathbf{Q}}
\newcommand{\K}{\mathbf{K}}
\newcommand{\Hb}{\mathcal{H}}
\newcommand{\F}{\mathcal{F}}
\newcommand{\G}{\mathcal{G}}
\newcommand{\R}{\mathbb{R}}
\newcommand{\Phib}{\mathbf{\Phi}}
\newcommand{\phib}{\bm{\phi}}
\newcommand{\rhob}{\bm{\rho}}
\newcommand{\J}{\mathcal{J}}

\newcommand{\onedot}{$\mathsurround0pt\ldotp$}
\newcommand{\eg}{\emph{e.g}\onedot} \newcommand{\Eg}{\emph{E.g}\onedot}
\newcommand{\ie}{\emph{i.e}\onedot} \newcommand{\Ie}{\emph{I.e}\onedot}
\newcommand{\cf}{\emph{c.f}\onedot} \newcommand{\Cf}{\emph{C.f}\onedot}
\newcommand{\etc}{\emph{etc}\onedot} \newcommand{\vs}{\emph{vs}\onedot}
\newcommand{\wrt}{w.r.t\onedot} \newcommand{\dof}{d.o.f\onedot}
\newcommand{\etal}{\emph{et al}\onedot }

\graphicspath{{./figures/}{./figure/images/}}
\IEEEtitleabstractindextext{%

\begin{abstract}
We study the problem of learning to rank from multiple information sources. Though multi-view learning and learning to rank have been studied extensively leading to a wide range of applications, multi-view learning to rank as a synergy of both topics has received little attention. 
The aim of the paper is to propose a composite ranking method while keeping a close correlation with the individual rankings simultaneously.
We present a generic framework for multi-view subspace learning to rank (MvSL2R), and two novel solutions are introduced under the framework.
The first solution captures information of feature mappings from within each view as well as across views using autoencoder-like networks.
Novel feature embedding methods are formulated in the optimization of multi-view unsupervised and discriminant autoencoders.
Moreover, we introduce an end-to-end solution to learning towards both the joint ranking objective and the individual rankings. The proposed solution enhances the joint ranking with minimum view-specific ranking loss, so that it can achieve the maximum global view agreements in a single optimization process.
The proposed method is evaluated on three different ranking problems, i.e. university ranking, multi-view lingual text ranking and image data ranking, 
providing superior results compared to related methods.
\end{abstract}

\begin{IEEEkeywords}
Learning to rank, multi-view data analysis, ranking
\end{IEEEkeywords}}
\maketitle

\IEEEdisplaynontitleabstractindextext

\ifCLASSOPTIONpeerreview
\begin{center} \bfseries EDICS Category: 3-BBND \end{center}
\fi
%
\IEEEpeerreviewmaketitle

\section{Introduction}
Learning to rank is an important research topic in information retrieval and data mining, which aims to learn a ranking model to produce a query-specfic ranking list. The ranking model establishes a relationship between each pair of data samples by combining the corresponding features in an optimal way \cite{Liu2009a}. A score is then assigned to each pair to evaluate its relevance forming a joint ranking list across all pairs. The success of learning to rank solutions has brought a wide spectrum of applications, including online advertising \cite{Zhu2009}, natural language processing \cite{Amini2009} and multimedia retrieval \cite{Yu2015}.\par
Learning appropriate data representation and a suitable scoring function are two vital steps in the ranking problem. Traditionally, a feature mapping models the data distribution in a latent space to match the relevance relationship, while the scoring function is used to quantify the relevance measure \cite{Liu2009a}; however, the ranking problem in the real world emerges from multiple facets and data patterns are mined from diverse domains. For example, universities are positioned differently based on numerous factors and weights used for quality evaluation by different ranking agencies. Therefore, a global agreement across sources and domains should be achieved while still maintaining a high ranking performance.\par
Multi-view learning has received a wide attention with a special focus on subspace learning \cite{Cao2017, Kan2016} and co-training \cite{Blum1998}, and few attempts have been made in ranking problems \cite{Usunier2011}. It introduces a new paradigm to jointly model and combine information encoded in multiple views to enhance the learning performance. Specifically, subspace learning finds a common space from different input modalities using an optimization criterion. 
Canonical Correlation Analysis (CCA) \cite{Hardoon2004, Rupnik2010} is one of the prevailing unsupervised method used to measure a cross-view correlation. 
By contrast, Multi-view Discriminant Analysis (MvDA) \cite{Kan2016} is a supervised learning technique seeking the most discriminant features across views by maximizing the between-class scatter while minimizing the within-class scatter in the underlying feature space.
Furthermore, a generalized multi-view embedding method \cite{Cao2017}
was proposed using a graph embedding framework for numerous unsupervised and supervised learning techniques with extension to nonlinear transforms including (approximate) kernel mappings \cite{Iosifidis2014, Iosifidis2016b} and neural networks \cite{Cao2017, Andrew2013}. A nonparametric version of \cite{Cao2017} was also proposed in \cite{Cao2017a}.
On the other hand, co-training \cite{Blum1998} was introduced to maximize the mutual agreement between two distinct views, and can be easily extended to multiple inputs by subsequently training over all pairs of views. A solution to the learning to rank problem was provided by minimizing the pairwise ranking difference using the same co-training mechanism \cite{Usunier2011}.\par
\begin{figure}
\centering
\includegraphics[width=.48\textwidth]{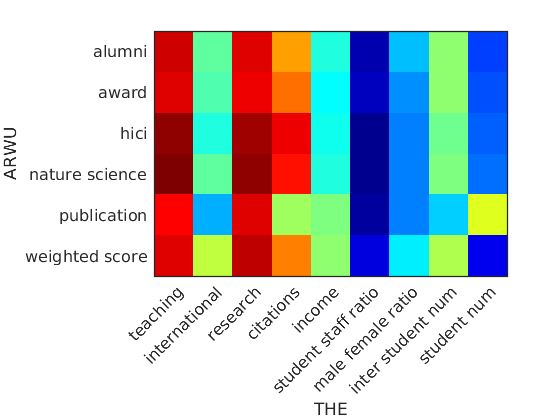}
\caption{The correlation matrix between the measurements of Times Higher Education (THE) and Academic Ranking of World Universities (ARWU) rankings. 
The data is extracted and aligned based on the performance of the common universities in 2015 between the two ranking agencies. The reddish color indicates high correlation, while the matrix elements with low correlation are represented in bluish colors. }\label{fig:corr_mat}
\end{figure}

Although there are several applications that could benefit from multi-view learning to rank approach, the topic has still been insufficiently studied up to date \cite{Feng2017}.
Ranking of multi-facet objects is generally performed using composite indicators. The usefulness of a composite indicator depends upon the selected functional form and the weights associated with the component facets. Existing solutions for university ranking are an example of using the subjective weights in the method of composite indicators. However, the functional form and its assigned weights are difficult to define. Consequently, there is a high disparity in the evaluation metric between agencies, and the produced ranking lists usually cause dissension in academic institutes. 
However, one observation is that, the indicators from different agencies may partially overlap and have a high correlation between each other. We present an example in Fig. \ref{fig:corr_mat} to show that, several attributes in the THE dataset \cite{THE2016}, including teaching, research, student staff ratio and student number are highly correlated with all of the attributes in the ARWU dataset \cite{Liu2005}. Therefore, the motivation of this paper is to find a composite ranking by exploiting the correlation between individual rankings. \par
Earlier success in multi-view subspace learning provides a promising way for composite ranking. Concatenating multiple views into a single input overlooks possible view discrepancy and does not fully exploit their mutual agreement in ranking. Our goal is to study beyond the concatenation of information from different sources used for ranking directly. This paper offers a multi-objective solution to ranking by capturing relevant information of feature mapping from within each view as well as across views. We propose a generic framework for multi-view subspace learning to rank (MvSL2R). It enables to incorporate novel feature embedding methods of both multi-view unsupervised and discriminant autoencoders.
Moreover, we propose an end-to-end method to optimize the trade-off between view-specific ranking and a discriminant combination of multi-view ranking. To this end, we can improve cross-view ranking performance while maintaining individual ranking objectives. \par
Intermediate feature representation in the neural network are exploited in our ranking solutions. 
Specifically, the first contribution is to provide two closely related methods by adopting an autoencoder-like network. We first train a network to learn view-specific feature mappings, and then maximize their correlation with the intermediate representations using either an unsupervised or discriminant projection to a common latent space. A stochastic optimization method is introduced to fit the correlation criterion. 
Both the autoencoding sub-network per view with a reconstruction objective and feedforward sub-networks with a joint correlation-based objective are iteratively optimized in the entire network. The projected feature representations in the common subspace are then combined and used to learn for the ranking function. \par
The second contribution (graphically described in Fig. \ref{fig:DMvDR}) is an end-to-end multi-view learning to rank solution. A sub-network for each view is trained with its own ranking objective. Then, features from intermediate layers of the sub-networks are combined after a discriminant mapping to a common space, and trained towards the joint ranking objective. As a result, a network assembly is developed to enhance the joint ranking with mimimum view-specific ranking loss, so that we can achieve the maximum view agreement within a single optimization process.\par
The rest of the paper is organized as follows. In Section \ref{sec:rel}, we describe the related work close to our proposed methods. The proposed methods are introduced in Section \ref{sec:model}. In Section \ref{sec:exp}, we present quantitative results to show the effectiveness of the proposed methods. Finally, Section \ref{sec:cl} concludes the paper.
\section{Related Work}\label{sec:rel}
\subsection{Learning to rank}
Learning to rank aims to optimize the combination of data representation for ranking problems \cite{liu2009}. It has been widely used in a number of applications, including image retrieval and ranking \cite{Yu2015, Li2017}, image quality ratings \cite{Wu2016}, online advertising \cite{Zhu2009}, and text summarization \cite{Usunier2011}. Solutions to this problem can be decomposed into several key components, including the input feature, the output vector and the scoring function. The framework is developed by training the scoring function from the input feature to the output ranking list, and then, scoring the ranking of new data.
Traditional methods also include engineering the feature using the PageRank model \cite{Page1999}, for example, to optimally combine them for obtaining the output. Later, research was focused on discriminatively training the scoring function to improve the ranking outputs. The ranking methods can be organized in three categories for the scoring function: the pointwise approach, the pairwise apporach, and the listwise approach. \par
We consider the pairwise approach in this paper and review the related methods as follows. 
A preference network is developed in \cite{Cohen1998} to evaluate the pairwise order between two documents. The network learns the preference function directly to the binary ranking output without using an additional scoring function. RankNet \cite{Burges2005} defines the cross-entropy loss and learns a neural network to model the ranking. 
Assuming the scoring function to be linear \cite{Herbrich2000}, the ranking problem can be transformed to a binary classification problem, and therefore, many classifiers are available to be applied for ranking document pairs. RankBoost \cite{Freund2003} adopts Adaboost algorithm \cite{Freund1995}, which iteratively focuses on the classfication errors between each pair of documents, and subsequently, improves the overall output. Ranking SVM \cite{Joachims2002} applies SVM to perform pairwise classification. GBRank is a ranking method based on Gradient Boost Tree \cite{Friedman2001}. Semi-supervised multi-view ranking (SmVR) \cite{Usunier2011} follows the co-training scheme to rank pairs of samples. Moreover, recent efforts focus on using the evaluation metric to guide the gradient with respect to the ranking pair during training. These studies include AdaRank \cite{Xu2007}, which optimizes the ranking errors rather than the classification error in an adaptive way, and LambdaRank \cite{Burges2007}. However, all of these methods above consider the case of single view inputs, while limited work on multi-view learning to rank has been studied \cite{Feng2017, Ye2015, Gao2014}.  
\subsubsection{Bipartite ranking}\label{sec:pair}
The pairwise approach of the ranking methods serves as the basis of our ranking method, and therefore, is reviewed explicitly in this section.
Suppose the training data is organized in query-sample pairs $\{(\x_i^q, \y_i^q)\}$, where $q \in \{1,2,\dots,Q\}$ is a query, $\x_i^q \in \mathrm{R}^d$ is the $d$-dimensional feature vector of the pair of query $q$ and its $i$-th retrieved sample, $\y^q_i \in \{0,1\}$ is their relevance score, and the number of query-specific samples is $N_q$. We perform the pairwise transformation before the relevance prediction of each query-sample pair, so that only the samples that belong to the same query are evaluated \cite{Herbrich2000}.\par 
The modeled probability between each pair in this paper is defined as
\begin{equation}
\mathbf{p}_{i}^q(\phi)=\frac{1}{1+\exp(\phi(\x_i) - \phi(\x_q))}, \nonumber
\end{equation}
where $\mathbf{\phi}:\x\rightarrow \R$ is the linear scoring function as $\phi(\x) = \mathbf{a}^\top \x$, which maps the input feature vectors to the scores. 
Due to its linearity, we can transform the feature vectors and relevance score into relative pairs as
\begin{equation}
(\x_m', \y_m') = (\x_q - \x_i, \y_i^q).\label{eq:rel}
\end{equation}
If we use an ordinal list $\mathbf{r}$ as the raw input, then the pairwise relevance is defined as 
\begin{equation}
\y^q_i =
  \begin{cases}
    1       & \quad \text{if } \mathbf{r}_i \leq \mathbf{r}_q\\
    0  & \quad \text{if }\mathbf{r}_i > \mathbf{r}_q.
\end{cases}
\end{equation}
Since the feature difference $\x'_m$ becomes the new feature vector as the input data for nonlinear transforms and subspace learning, the ranking probability is defined as 
\begin{equation}
    \mathbf{p}_{m}(\phi)=\frac{1}{1+\exp(-\phi(\x'_m))}
    =\frac{1}{1+\exp(-\mathbf{a}^\top \x'_m)}. \label{eq:proba} 
\end{equation}
The objective to make the right order of ranking can then be formulated as the cross entropy loss below,
\begin{align}
\ell_\text{Rank} &= \arg\min \sum_{q=1}^Q \sum\limits_{i=1}^{N_q}
\big(\y_i^q \log \mathbf{p}_i^q)+ (1-\y_i^q) \log \mathbf{p}_i^q)\big) \nonumber \\
&= \arg\min \sum_{m=1}^M
\big(\y'_m \log \mathbf{p}_m)+ (1-\y'_m) \log \mathbf{p}_m)\big), \label{eq:rank}
\end{align}%
which is proved in \cite{Burges2005} that it is an upper bound of the pairwise 0-1 loss function and optimized using gradient descent. The logistic regression or softmax function in neural networks can be used to learn the scoring function.

\subsection{Multi-view deep learning}
Multi-view learning considers enhancing the feature discriminability by taking inputs from diverse sources. One important approach to follow is subspace learning, which is traced back to CCA \cite{Hotelling1936,Borga2001} between two input domains, and its multi-view extension, which has been studied in \cite{Nielsen2002, Gong2014, Luo2015}. This approach can also be generalized using a higher-order correlation \cite{Luo2015}. The main idea behind these techniques is to project the data representations in the two domains to a common subspace optimizing their mutual correlation. Subspace learning with supervision has also been extensively studied. Multi-view Discriminant Analysis \cite{Kan2016} performs the dimensionality reducation of features from multiple views exploiting the class information. Recently, these methods were generalized in the same framework \cite{Cao2017, Sharma2012}, which accommodates multiple views, supervision and nonlinearity. Multi-view subspace clustering methods \cite{Zhang2018, Zhan2018, Abavisani2018} also provide ways to recover the low-dimensional data structure for clustering, while our problem is cast to a binary classification problem after pairwise transform. Co-training \cite{Blum1998} first trains two seperate regressors and then, iteratively maximizes their agreements. \par
Deep learning, which exploits the nonlinear transform of the raw feature space, has also been studied in the multi-view scenario. The multi-modal deep autoencoder \cite{Ngiam2011} was proposed by taking nonlinear representations of a pair of views to learn their common characteristics. Deep CCA \cite{Andrew2013} is another \emph{two-view} method which maximizes the pairwise correlation using neural networks. 
Thereafter, a two-view correlated autoencoder was developed \cite{Wang2015, Chandar2016} with objectives to correlate the view pairs but also reconstruct the individual view in the same network. By contrast, we propose a generic framework which is extensible for multiple views and both unsupervised and discriminant autoencoders for ranking. The previous work on discriminant autoencoders introduced an additional regularization term \cite{Nousi2017}, while our method embeds the class discrimination in the Laplacian matrix, and is extensible for multiple views. \par
Multi-view Deep Network \cite{Kan2016a} was also proposed as a neural network extension of MvDA \cite{Kan2016}. It optimizes the ratio trace of the graph embedding \cite{Yan2007} to avoid the complexity of solutions without a closed form \cite{Jia2009}. In this paper, however, we show that the trace ratio optimization can be solved efficiently in the updates of the multi-view networks. Moreover, multiple objectives are optimized in different layers and sub-network in our solution. Deep Multi-view Canonical Correlation Analysis (DMvCCA) and Deep Multi-view Modular Discriminant Analysis (DMvMDA) \cite{Cao2017} are closely related to our work, and hence, they are described in the following sections.
\subsubsection{Deep Multi-view Canonical Correlation Analysis (DMvCCA)}
\begin{singlespace}
The idea behind DMvCCA \cite{Cao2017} is to find a common subspace using a set of linear transforms $\W=[\W_1,\W_2,\dots,\W_v]^\top$ to project nonlinearly mapped input samples $\Z_v$ from the $v$th view where the correlation is maximized. Specifically, it aims to maximize
\end{singlespace}
{\footnotesize
\begin{equation}
\J_\text{DMvCCA}=\underset{\sum \limits_{i=1}^V \W_i^\top \Z_i\mathbf{L}\Z_i^\top \W_i=\I}{\arg\max} \Tr\bigg({\sum\limits_{i=1}^V \sum\limits_{\substack{j=1\\j\neq i}}^V} \W_i^\top \Z_i\, \mathbf{L}\,\Z_j^\top \W_j\bigg) 
\end{equation}}\label{eq:dmvcca}%
where the matrix $\mathbf{L}=\I-{1\over N}\e\e^\top$ centralizes the input data matrix of each view $v$, and $\e$ is a vector of ones, .
By defining the cross-view covariance matrix between views $i$ and $j$ as
$\mathbf{\Sigma}_{ij} = {1 \over N} \tilde{\Z}_i \tilde{\Z}_j$,
where $\tilde{\Z}_v, v=1,\dots,V$, is the centered view, the data projection matrix $\W$, which has the column vector of $\W_v$ in the $v$th view, can be obtained by solving the generalized eigenvalue problem
\begin{equation}
\begin{bmatrix}
\mathbf{0} &\mathbf{\Sigma}_{12}&\cdots & \mathbf{\Sigma}_{1V}\\[1mm]
\mathbf{\Sigma}_{21} &\mathbf{0}&\cdots & \mathbf{\Sigma}_{2V} \\[1mm]
\vdots&\vdots &\ddots & \vdots \\[1mm]
\mathbf{\Sigma}_{V1}&\mathbf{\Sigma}_{V2}&\cdots &  \mathbf{0}
\end{bmatrix}
\W
= \lambda \begin{bmatrix}
\mathbf{\Sigma}_{11}& \mathbf{0}&\cdots & \mathbf{0} \\[1mm]
\mathbf{0} &\mathbf{\Sigma}_{22} &\cdots & \mathbf{0} \\[1mm]
\vdots&\vdots &\ddots & \vdots \\[1mm]
\mathbf{0}&\mathbf{0}&\cdots &  \mathbf{\Sigma}_{VV}
\end{bmatrix} \W.
\end{equation}
It shows that the solution to this problem is derived with the maximal inter-view covariances and the minimal intra-view covariances.
\subsubsection{Deep Multi-view Modular Discriminant Analysis (DMvMDA)}
\begin{singlespace}
DMvMDA \cite{Cao2017} is the neural network-based multi-view solution of LDA which maximizes the ratio of the determinant of the between-class scatter matrix of all view pairs to that of the within-class scatter matrix. Mathematically, it is written as the projection matrix of the DMvMDA and is derived by optimizing function
\end{singlespace}
{\footnotesize
\begin{equation}
\J_{\text{DMvMDA}}= \underset{\sum \limits_{i=1}^V \W_i^\top \Z_i\mathbf{L}_W\Z_i^\top \W_i=\I}{\arg\max}
\Tr \big(\sum\limits_{i=1}^V\sum \limits_{j=1}^V \W_i^\top{\Z}_i \mathbf{L}_B \Z_j^\top \W_j
\big),
\end{equation}}%
where the between-class Laplacian matrix is
\begin{equation}
\mathbf{L}_B = 2 \, \sum_{p=1}^{C}\sum_{q=1}^{C}({1 \over N_p^2} \e_p \, \e_p^\top - {1 \over N_p N_q} \e_p\,\e_q^\top).\nonumber
\end{equation}
and the within-class Laplacian matrix is
\begin{equation}
\mathbf{L}_W=\I-\sum \limits_{c=1}^{C} {1 \over N_c} \e_c{\e_c}^\top.\nonumber
\end{equation}
\section{Model Formulation}\label{sec:model}
We first introduce the generic framework for multi-view subspace learning to rank (MvSL2R). It is followed by the formulations of MvCCAE and MvMDAE. Finally, the end-to-end ranking method is presented.
\subsection{Multi-view Subspace Learning to Rank (MvSL2R)}\label{sec:mvsl2r}
Multi-view subspace learning to rank is formulated based on the fact that the projected feature in the common subspace can be used to train a scoring function for joint ranking. We generate the training data from the intersection of ranking samples between views to have the same samples but various representations from different view origins. 
The overall ranking agreement is made by calculating the average voting from the intersected ranking orders as
\begin{equation}
\overline{\mathbf{r}} = {1 \over V}\sum_{v=1}^V \mathbf{r}_v.\label{eq:rbar}
\end{equation}
By performing the pairwise transform in section \ref{sec:pair} over the ranking data, we have the input feature $\mathcal{X}=\{\X_1, \X_2,\dots,\X_V\}$ and $\mathcal{Y}=\{\y_1,\y_2,\dots,\y_V\}$ of $V$ views. The cross-view relevance scores $\overline{\y}$ is also obtained from the average ranking orders $\overline{\mathbf{r}}$ using \eqref{eq:rel}. 
The proposed ranking method consists of feature mapping into a common subspace, training a logistic regressor as the scoring function, and predicting the relevance of new sample pairs using the probability function
\begin{equation}
\mathbf{p}_v(\X_v)=\frac{1}{1+\exp(-\mathbf{a}^\top\W_v^\top\F_v(\X_v))}, \label{eq:pred}
\end{equation}
where $\W_v$ is the data projection matrix of the $v$th view, and $\mathbf{a}$ is the weight from the logistic regressor described in \eqref{eq:proba}. Let us denote time $T$ as the number of epochs over the training data, and then we can summarize the learning steps in the algorithm at time $t$ below.
\begin{algorithm}[h!]
\SetKwInOut{Input}{Input}
\SetKwInOut{Output}{Output}
\SetKwInOut{Init}{Init}
\SetKwInOut{Train}{Train}
\SetKwInOut{Predict}{Predict}
\underline{Framework MvSL2R} $(\mathcal{X},\mathcal{R},k)$\;
\Input{The feature vectors of V views $\mathcal{X}=\{\X_1,\X_2,\dots,\X_V\}$, the ranking orders of V views $\mathcal{R}=\{\mathbf{r}_1,\mathbf{r}_2,\dots,\mathbf{r}_V\}$, and the dimensionality in the subspace $k$.}
\Output{The joint relevance probabilities $\overline{\mathbf{p}}$ of the new data.}
\Init \\
Perform pairwise transform in \eqref{eq:rel} to obtain relative samples and relevance $\y_v$ of each view, and their cross-view relevance scores $\overline{\y}$ from the average ranking $\overline{\mathbf{r}}$ using \eqref{eq:rbar}.\\
\Train\\
\For {$t=1:T$}
{Update the sub-network $\G_v$ to minimize the view-specific reconstruction errors in MvCCAE and MvMDAE, or minimize the ranking loss using \eqref{eq:rank_loss} in DMvDR.\\
Update multi-view subspace embedding layer using \eqref{eq:mvccae_update} in MvCCAE or \eqref{eq:mvmdae_update} in MvMDAE and DMvDR. \\
The fused ranking loss is integrated in $\Hb$ in DMvDR and optimized in \eqref{eq:rank_loss}.\\
The encoding sub-network $\F_v$ in MvCCAE and MvMDAE is updated according to the chain rule, while the gradient of view-specific sub-network in DMvDR is defined in \eqref{eq:dmvdr_f}.}
In MvCCAE and MvMDAE, we learn the separate linear weight $\mathbf{a}$ in the scoring function \eqref{eq:proba} from the output of the joint embedding layer to minimize the ranking error to cross-view relevance scores $\overline{\y}$.\\
\Predict\\
Predict the relevance probabilities of the new sample pairs using \eqref{eq:pred} in MvCCAE and MvMDAE, while DMvDR predicts $\overline{\mathbf{p}}$ through the view-specific $\F_v$ and fused sub-network $\Hb$.
\caption{Multi-view Subspace Learning to Rank.}\label{algo}
\end{algorithm}
\subsection{Multi-view Canonically Correlated Auto-Encoder (MvCCAE)}
\begin{singlespace}
In contrast to DMvCCA and DMvMDA, where the nonlinear correlation between multiple views is optimized, we propose a multi-objective solution by maximizing the between-view correlation while minimizing the reconstruction error from each view source. Given the data matrix $\mathcal{X}=\{\X_1,\X_2,\dots,\X_V\}$ of $V$ views, the encoding network $\F$ and the decoding network $\G$, and the projection matrix $\W$, the objective of MvCCAE is formulated as follows,
\end{singlespace}
{\small
\begin{equation}
\J_{\text{MvCCAE}} = \arg \max \, \J'_{\text{DMvCCA}} - \alpha \sum_v^V \, 
\ell_\text{AE}\Big (\X_v; \G_v(\F_v(\cdot))\Big ),\label{eq:mvccae}
\end{equation}}%
where we introduce the new objective 
\begin{equation}
\J'_{\text{DMvCCA}} = \underset{\W^\top \W = \I}{\arg\max} 
\frac{ \Tr\bigg({\sum\limits_{i=1}^V \sum\limits_{\substack{j=1\\j\neq i}}^V} \W_i^\top \Z_i\, \mathbf{L}\,\Z_j^\top \W_j\bigg)}
{\Tr\big(\sum \limits_{i=1}^V  \W_i^\top \Z_i\, \mathbf{L}\,\Z_i^\top \W_i\big)}, \label{eq:dmvcca2}
\end{equation}
and the loss function of the $v$th autoencoder is $\ell_\text{AE}(\X_v; \G_v(\F_v(\cdot)))=\|\X_v-\G_v(\F_v(\X_v))\|_2 + \rho \sum_l\|\nabla_{\X_v} \F_v^l(\X_v)\|_2$, with the $L_2$ regularization at the $l$th intermediate layer of the $v$th view denoted by $\Z_v^l=\F_v^l(\X_v)$. Here, $\alpha$ and $\rho$ are the controlling parameters for the regularization.
\subsubsection{Optimization}
Following the objective of DMvCCA \cite{Cao2017}, we aim to directly optimize the trace ratio in \eqref{eq:dmvcca2} and let 
\begin{equation}
    f=\operatorname{Tr}\Bigg(\,{\sum\limits_{i=1}^V \sum\limits_{\substack{j\neq i\\j=1}}^V} \W_i^\top \Z_i\, {\mathbf{L}}\,\Z_j^\top \W_j \Bigg),\label{eq:f1}
\end{equation}
and 
\begin{equation}
g=\operatorname{Tr}\Bigg(\,{\sum\limits_{i=1}^V \W_i^\top \Z_i\, {\mathbf{L}}\,\Z_i^\top \W_i \Bigg)}.\label{eq:g1}
\end{equation}
Here, the output of each sub-network $\F_v$ is denoted by $\Z_v=\F_v(\X_v)$. Then, we have 
\begin{equation}
\frac{\partial f}{\partial \Z_i}
={\sum\limits_{i=1}^V \sum\limits_{\substack{j\neq i\\j=1}}^V}
\W_i \,\W_j^\top \Z_j \,\mathbf{L},\label{eq:fg1}
\end{equation}
and 
\begin{equation}
\frac{\partial g}{\partial \Z_i}
={\sum\limits_{i=1}^V} 
\W_i \,\W_i^\top \Z_i \,\mathbf{L}.\label{eq:gg1}
\end{equation}
\begin{singlespace}
By using \eqref{eq:f1}, \eqref{eq:g1}, \eqref{eq:fg1} and \eqref{eq:gg1}, and following the quotient rule, we derive the gradient of the new DMvCCA w.r.t the weighted input $\Z_v$ to the common layer as
\end{singlespace}
{\footnotesize
\begin{equation}
\frac{\partial  \J'_{\text{DMvCCA}}}{\partial \Z_v}=\frac{1}{g^2}  \bigg(g\, \frac{\partial f}{\partial \Z_v} -f \, \frac{\partial g}{\partial \Z_v}\bigg),
\end{equation}}%
and the update of the common layer is
{\footnotesize
\begin{equation}
\frac{\partial \J_{\text{MvCCAE}}}{\partial \Z_v}=
\frac{\partial \J'_{\text{DMvCCA}}}{\partial \Z_v}-\alpha \, \frac{\partial \ell_\text{AE}}{\partial \Z_v}. \label{eq:mvccae_update}
\end{equation}}%
The sub-networks $\F_v$ and $\G_v$ are optimized using the gradient with respect to their network parameters which can be easily derived using the chain rule depending on its activation functions.
\subsection{Multi-view Modularly Discriminant Auto-Encoder (MvMDAE)}
\begin{singlespace}
Similar to MvCCAE, the objective of MvMDAE is to optimize the combination of the view-specific reconstruction error and the cross-view correlation as follows,
\end{singlespace}
{\footnotesize
\begin{equation}
\J_{\text{MvMDAE}} = \arg \max \J'_{\text{DMvMDA}} - \alpha \sum_v^V \, 
    \ell_\text{AE}\Big (\X_v;  \G_v(\F_v(\cdot))\Big).\label{eq:mvmdae}
\end{equation}}%
The new objective for the cross-view correlation is
\begin{equation}
\J'_{\text{DMvMDA}}= \underset{\W^\top \W=\I}{\arg\max}
\frac{\Tr \big(\sum\limits_{i=1}^V\sum \limits_{j=1}^V \W_i^\top{\Z}_i \mathbf{L}_B \Z_j^\top \W_j
\big)}
{\Tr\big(\sum \limits_{i=1}^V \W_i^\top \Z_i \mathbf{L}_W  \Z_i^\top \W_i\big)},\label{eq:dmvmda2}
\end{equation}
\subsubsection{Optimization}
The detailed optimization is derived by replacing the laplacian matrix in MvCCAE with $\mathbf{L}_B$ and $\mathbf{L}_W$ in \eqref{eq:dmvmda2}.
We let 
\begin{equation}
    f=\operatorname{Tr}\Bigg(\,{\sum\limits_{i=1}^V \sum\limits_{\substack{j\neq i\\j=1}}^V} \W_i^\top \Z_i\, {\mathbf{L}_B}\,\Z_j^\top \W_j \Bigg),\label{eq:f2}
\end{equation}
and 
\begin{equation}
g=\operatorname{Tr}\Bigg(\,{\sum\limits_{i=1}^V \W_i^\top \Z_i\, {\mathbf{L}_W}\,\Z_i^\top \W_i \Bigg)}.\label{eq:g2}
\end{equation}
Then, we obtain the gradient of $f$ and $g$ as follows, 
\begin{equation}
\frac{\partial f}{\partial \Z_i}
={\sum\limits_{i=1}^V \sum\limits_{\substack{j\neq i\\j=1}}^V}
\W_i \,\W_j^\top \Z_j \,\mathbf{L}_B,\label{eq:fg2}
\end{equation}
and
\begin{equation}
\frac{\partial g}{\partial \Z_i}
={\sum\limits_{i=1}^V} 
\W_i \,\W_i^\top \Z_i \,\mathbf{L}_W.\label{eq:gg2}
\end{equation}%
\noindent The gradient of the new DMvMDA can be derived by using \eqref{eq:f2}, \eqref{eq:g2}, \eqref{eq:fg2} and \eqref{eq:gg2}, and applying the quotient rule as follows,
\begin{equation}
    \frac{\partial \J'_{\text{DMvMDA}}}{\partial \Z_v}=\frac{1}{g^2}  \bigg(g\, \frac{\partial f}{\partial \Z_v} -f \, \frac{\partial g}{\partial \Z_v}\bigg).
\end{equation}
The gradient of the multi-view subspace embedding layer is
\begin{equation}
\frac{\partial \J_{\text{MvMDAE}}}{\partial \Z_v}=
\frac{\partial \J'_{\text{DMvMDA}}}{\partial \Z_v}-\alpha \, \frac{\partial \ell_\text{AE}}{\partial \Z_v}\label{eq:mvmdae_update}
\end{equation}
Using the chain rule, we update encoder $\F_v$ and decoder $\G_v$ depending its activation functions. 
\begin{figure*}
\centering
\includegraphics[width=.8\textwidth]{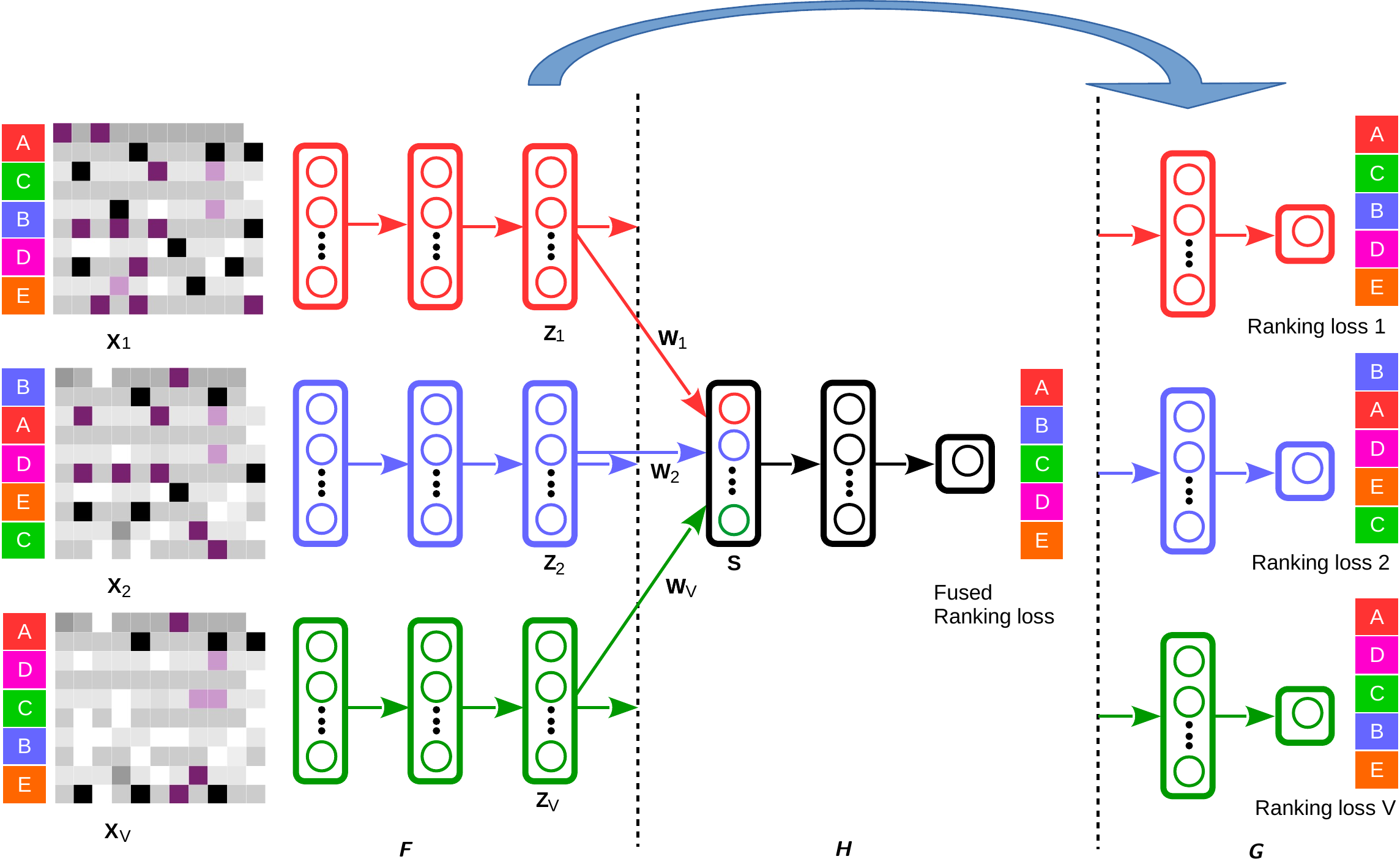}
\caption{System diagram of the Deep Multi-view Discriminant Ranking (DMvDR). First, the features $\mathcal{X}=\{\X_1,\X_2,\dots,\X_V\}$ are extracted for data representations in different views and fed through the individual sub-network $\F_v$ to obtain the nonlinear representation $\Z_v $ of the $v$th view. The results are then passed through two pipelines of networks. 
One line goes to the projection $\W$, which maps all $\Z_v$ to the common subspace, and their concatenation is trained to optimize the fused ranking loss with the fused sub-network $\Hb$. The other line connects $\Z_v$ to the sub-network $\G_v, \forall v=1,\dots,V$ for the optimization of the $v$th ranking loss. 
}\label{fig:DMvDR}
\end{figure*}
\subsection{Deep Multi-view Discriminant Ranking (DMvDR)}\label{sec:dmvdr}
Multi-view Subspace Learning to Rank provides a promising method with MvCCAE and MvMDAE. However, they do not have direct connections to each view-specific ranking. To improve the global agreement between views by following the same framework, we propose to optimize the view-specific and the joint ranking together in an integrated neural network as shown in Fig. \ref{fig:DMvDR}. Taking university ranking as an example, various ranking lists are generated from different agencies, and each agency uses a different set of attributes to represent the universities.
During training, given the inputs $\mathcal{X}=\{\X_1,\X_2,\dots,\X_V\}$, the cross entropy loss \eqref{eq:rank} is optimized against the view-specific relevance $\y_v$ and the joint view relevance $\overline{\y}$. Based on different ranking losses, the attributes $\X_v, v=1,\dots,V$, are trained through the view-specific sub-network $\F_v$. 
The nonlinear representations $\Z_v=\F_v(\X_v), v=1,\dots,V$, are the inputs of the joint network $\mathcal{H}$ as $\W_v^\top \Z_v$, $v=1,\dots,V$, before the mappings to generate the joint university ranking list. 
$\Z_v$ is also the input to the view-specific network $\G_v$, which minimizes its distance to the respective relevance $\y_v$. We exploit the effectiveness of intermediate layers $\Z_v$ through the view-specific sub-networks $\F_v$ and $\G_v$ towards the view-specific ranking loss in DMvDR.
The detailed procedure of this method is described below.\par
The gradient of the ranking loss of each view ${\partial \ell_{\text{rank}}\over \partial \mathbf{p}_v}$ and fused ranking loss across views ${\partial \ell_{\text{rank}}\over \partial \overline{\mathbf{p}}}$ can be derived as 
\begin{equation}
    \frac{\partial \ell_{\text{rank}}}{\partial \mathbf{p}_i}={1\over N}\sum_{i=1}^{N}\frac{\mathbf{p}_i-\y_i}{(1-\mathbf{p}_i)\mathbf{p}_i},\label{eq:rank_loss}
\end{equation}
where $\mathbf{p}_i$ denotes the ranking probability of the $i$-th sample in either $\mathbf{p}_v$ or $\overline{\mathbf{p}}$.
We update the view-specific sub-network $\G_v$ from the input $\Z_v$ and the output $\mathbf{p}_v$. Since it passes through the view-specific $\F_v$ and $\G_v$ sub-network, $\partial \mathbf{p}_v\over\partial \G_v$ and $\partial \mathbf{p}_v\over\partial \Z_v$ can be determined based on the activation functions and update $\G_v$ through backpropagation \cite{LeCun1998a}. \par
The fused sub-network $\mathcal{H}$ is updated with the gradient of the ranking loss from the cross-view relevance scores $\overline{\y}$.
In order to obtain $\overline{\y}$, we find the intersection of the ranking data with different attributes from various sources, and perform the pairwise transform to have the sample pairs as the input $\mathcal{X}$ and $\overline{\y}$ from the cross-view ranking orders $\overline{\mathbf{r}}$ in \eqref{eq:rbar}. 
As a result, the input $\mathbf{S}$ to the fused sub-network $\Hb$ is the concatenation of the nonlinear mappings from $V$ view-specific networks $\F_v$ as 
\begin{equation}
    \mathbf{S}=[\W^\top_1\Z_1\;\; \W^\top_2\Z_2\; \dots \;\W^\top_V\Z_V]^\top.\label{eq:s} 
\end{equation}
For testing, we can distinguish two possible scenarios: (a) If the samples are aligned and all presented from each view, the results from nonlinear mappings are combined in the same manner as the training phase to generate a fused ranking list $\overline{\mathbf{p}}$ at the end of $\Hb$ sub-network; and (b) If there are missing samples or completely unaligned in the test data, $\mathbf{S}=\W^\top_v \Z_v$ for the $v$th view. The resulting view-specific prediction $\mathbf{p}_v$ still maintains the cross-view agreement which is ranked from the trained joint network. 
The gradient of $\partial \overline{\mathbf{p}} \over \partial \Z_v$ and $\partial \overline{\mathbf{p}} \over \partial \Hb$ can be computed to update $\mathcal{H}$. \par
Joint ranking is achieved using a multi-view subspace embedding layer same as the one in MvMDAE, and its gradient is computed using \eqref{eq:mvmdae}.
The embedding layer is important as its gradient is forward passed to the fused sub-network $\Hb$. Meanwhile, it is backward propagated in the layers of $\F_v$ to reach the input $\X_v$. In turn, the parameters in $\G_v$ are also affected by the outputs of $\F_v$ sub-networks.
\par
The update of the view-specific $\F_v$ depends on both the view-specific ranking loss and the cross-view ranking loss as it is a common sub-network for both objectives. Through backpropagation, the $v$-th sub-networks $\F_v$ and $\G_v$ are optimized consecutively using gradients $\partial \ell_\text{Rank} \over \partial \F_v$ and $\partial \ell_\text{Rank} \over \partial \G_v$. Meanwhile, the training error comparing to the joint ranking $\overline{\y}$ is passed through multi-view subspace embedding layer $\mathbf{S}$ in \eqref{eq:s} as the input to the fused sub-network $\Hb$. The resulting gradient of each sub-network $\F_v$ is given by 
\begin{align}
{\partial \J_{\text{DMvDR}} \over \partial \F_v}= & {\partial \J_{\text{MvMDA}}\over \partial \F_v} 
- \alpha \, {\partial \over \partial \F_v} \ell_\text{Rank}(\X_v, \y_v; \G_v(\F_v(\cdot))) \nonumber \\
& - \beta \, {\partial \over \partial \F_v}\ell_\text{Rank}(\mathbf{S}, \overline{\y}; \Hb(\cdot)),\label{eq:dmvdr_f}
\end{align}
where $\alpha$ and $\beta$ are the scaling factors controlling the magnitude of the ranking loss. Similar to the other sub-networks, the gradients with respect to their network parameters can be obtained by following the chain rule from other partial derivatives.\par
The update of the entire network of DMvDR can be summarized using the SGD with mini-batches. The parameters of the sub-network are denoted by
$\theta=\{\theta_{\F_1},\theta_{\F_2},\dots,\theta_{\F_V},  \theta_{\G_1},\theta_{\G_2},\dots,\theta_{\G_V},\theta_\Hb\}$. A gradient descent step is $\Delta \theta = -\eta {\partial \over \partial \theta}\J_{\text{DMvDR}}$, where $\eta$ is the learning rate. The gradient update step at time $t$ can be written down with the chain rule collectively:
\begin{align}
\Delta\theta^t\;\;  =& \,-\eta\cdot\{\Delta\theta^t _{\F_1}, \Delta\theta^t_{\F_2},\dots,
\Delta \theta^t_{\F_V}, \nonumber \\ 
& \,\,\,\,\Delta \theta^t_{\G_1},\Delta \theta^t_{\G_2},\dots,\Delta \theta^t_{\G_V},\Delta \theta^t_\Hb\} \nonumber \\[1mm]
\Delta \theta^t_{\G_v} = & -\frac{\partial \ell_{\text{rank}}}{\partial \mathbf{p}^t_v} \cdot \frac{\partial \mathbf{p}^t_v}{\partial \G^t_v} \nonumber \\[1mm]
\Delta \theta^t_{\Hb} \, = & -\frac{\partial \ell_{\text{rank}}}{\partial \overline{\mathbf{p}}^t} \cdot \frac{\partial \overline{\mathbf{p}}^t}{\partial \Hb^t} \nonumber \\[1mm] 
\Delta \theta^t_{\F_v} = & 
\frac{\partial \J_{\text{MvMDA}}}{\partial \Z^t_v} \cdot \frac{\partial \Z^t_v}{\partial \F^t_v} 
-\alpha \, \frac{\partial \ell_{\text{rank}}}{\partial \mathbf{p}^t_v} \cdot \frac{\partial \mathbf{p}^t_v}{\partial \Z^t_v} \cdot \frac{\partial \Z^t_v}{\partial \F^t_v} \nonumber \\
& - \beta \, \frac{\partial \ell_{\text{rank}}}{\partial \overline{\mathbf{p}}^t} \cdot \frac{\partial \overline{\mathbf{p}}^t}{\partial \mathbf{S}^t} \cdot \frac{\partial \mathbf{S}^t}{\partial \Z^t_v} \cdot \frac{\partial \Z^t_v}{\partial \F^t_v}.
\end{align}\par
We generate the training data using the pairwise transform presented in Section \ref{sec:pair}. The weights are normalized to the unit norm during backpropagation.
In testing, the test samples can either be transformed into pairs firstly to evaluate the relative relevance of each sample to its query, or directly fed into the trained model to predict their overall ranking positions.
\section{Experiments}\label{sec:exp}
In this section, we evaluate the performance of the proposed multi-view learning to rank methods in three challenging problems: university ranking, multi-linguistic ranking and image data ranking. The proposed methods are also compared to the related subspace learning and co-training methods. The subspace learning methods follow the steps proposed in Section \ref{sec:mvsl2r} for ranking. All neural network topologies are chosen based on a validation set and trained for $100$ epoches. We compare the performance of the following methods in the experiments: 
\begin{itemize}
    \item \textbf{Best Single View}: a method which shows the best performance of Ranking SVM \cite{Joachims2002} over the individual views.
\item \textbf{Feature Concat}: a method which concatenate the features of the common samples for training a Ranking SVM \cite{Joachims2002}.
\item \textbf{LMvCCA \cite{Cao2017}}: a linear multi-view CCA method. 
\item \textbf{LMvMDA \cite{Cao2017}}: a linear supervised method for multi-view subspace learning. 
\item \textbf{MvDA \cite{Kan2016}}: another linear supervised method for multi-view subspace learning. It differs from the above in that the view difference is not encoded in this method. 
\item \textbf{SmVR \cite{Usunier2011}}: a semi-supervised method that seeks a global agreement in ranking. It belongs to the category of co-training. We develop the complete data in the following experiments for training so that its comparison with the subspace learning methods is fair. Therefore, SmVR becomes a supervised method in this paper. 
\item \textbf{DMvCCA \cite{Cao2017}}: a nonlinear extension of LMvCCA using neural networks.
\item \textbf{DMvMDA \cite{Cao2017}}: a nonlinear extension of LMvMDA using neural networks.
\item \textbf{MvCCAE}: the first proposed multi-view subspace learning to rank method proposed in the paper. 
\item \textbf{MvMDAE}: the supervised multi-view subspace learning to rank method proposed in the paper. 
\item \textbf{DMvDR}: the end-to-end multi-view learning to rank method proposed in the paper. 
\end{itemize}
We present the quantitative results using several evaluation metrics including the Mean Average Precision (MAP), classification accuracy and Kendal's tau. 
The Average Precision (AP) measures the relevance of all query and sample pairs with respect to the same query, while the MAP score calculates the mean AP across all queries \cite{Manning2008}. After performing pairwise transform on the ranking data, the relevance prediction can be considered as a binary classification problem, and therefore the classification is utilized for evaluation. Kendal's tau measures the ordinal association between two lists of samples.\par 
We also present the experimental results graphically, and the following measures are used. The {Mean Average Precision} (MAP) score, which is the average precision at the ranks where recall changes, is illustrated on the 11-point interpolated precision-recall curves (PR curve) to show the ranking performance. Also, the ROC curve provides a graphical representation of the binary classification performance. It shows the true positive rates against the false positive rate at different thresholds. The correlation plots show linear correlation coefficients between two ranking lists.\par 
\subsection{University Ranking}
\begin{figure}
[h!]
\begin{subfigure}[b]{.49\linewidth}
\centering
\includegraphics[width=.8\textwidth, height=3cm]{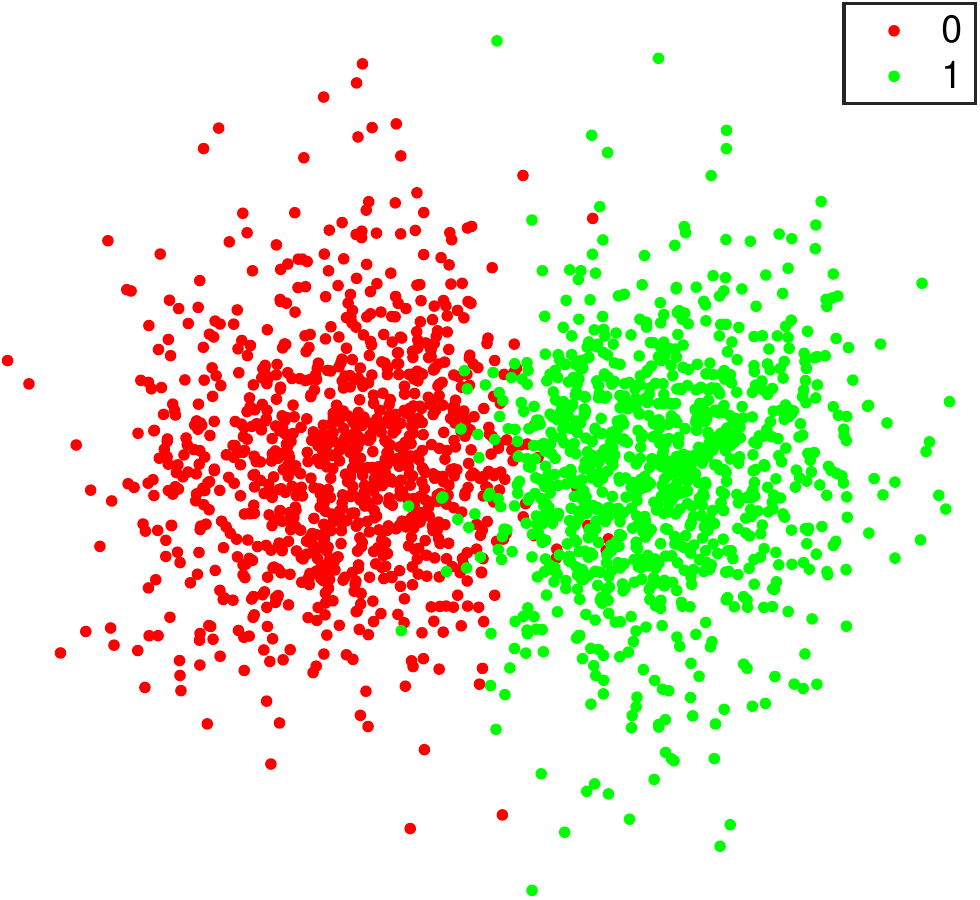}
\caption{Raw data distribution.}
\end{subfigure}
\begin{subfigure}[b]{.49\linewidth}
\centering
\includegraphics[width=.8\textwidth, height=3cm]{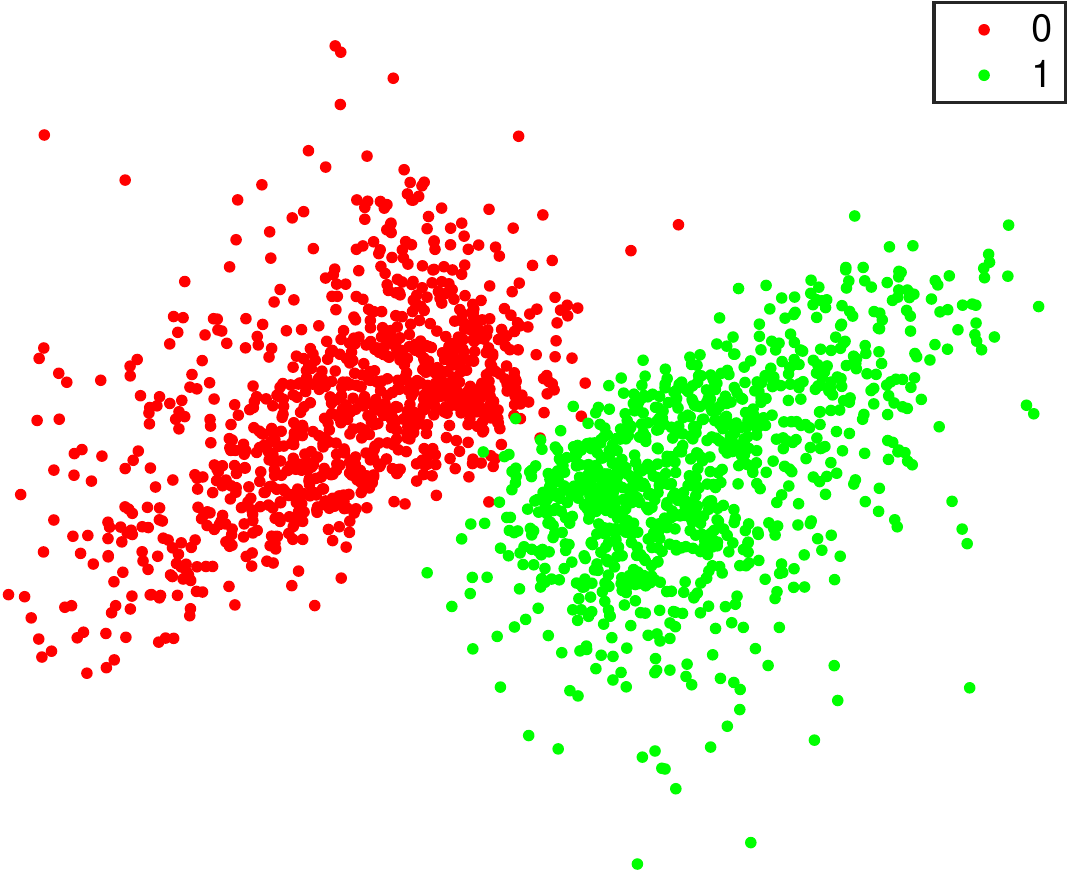}
\caption{Projected data distribution.}
\end{subfigure}
\caption{The left plot shows the data distribution by concatenating the measurements as features of the common universities from $3$ different agencies in 2015. The right plot shows the concatenated and projected features using MvMDAE for the same universities.}\label{fig:dist} 
\end{figure}

The university ranking dataset available in Kaggle.com \cite{kaggle2016} collects the world ranking data from three rating agencies, including the Times Higher Education (THE) World University Ranking, the  Academic Ranking of World Universities (ARWU), and the Center for World University Rankings (CWUR). Despite political and controversial influences, they are widely considered as authorities for university ranking. The measurements are used as the feature vectors after feature preprocessings, which includes feature standardization and removal of categorical variables and groundtruth indicators including the ranking orders, university name, location, year and total scores.
The $271$ common universities from 2012 to 2014 are considered for training. After the pairwise transform in each year, $36 542$ samples are generated as the training data. The entire data in 2015 is considered for testing. The data distribution (after binary transform) of the $196$ common universities in 2015 is shown in Fig. \ref{fig:dist}. 
We used a topology of two hidden layers, having $16$ and $32$ neurons each, and 10 output neurons for the encoding networks $\F_v$ for both MvCCAE and MvMDAE. The decoding network of each view has 64 hidden neurons before reconstructing to the input. Sigmoid activation function is used for all neurons in the networks. We used a minibatch size of 200 samples, the learning rate is $10^{-6}$ and $\rho$ for $L_2$ penalty is set to $10^{-9}$. In DMvDR, the encoding networks $\F_v$ consist of two hidden layers with $50$ and $10$ hidden neurons each and one output neuron, and the decoding networks $\G_v$ have $100$ hidden neurons. 
The joint layer $\Hb$ has $100$ neurons too. We used a minibatch size of $200$ samples, the learning rate is $10^{-4}$ and $\rho$ for $L_2$ penalty is set to $10^{-2}$.
\par 
We can make several observations from the data distribution in Fig. \ref{fig:dist}. Firstly, the pairwise transform is applied on the university ranking data, which equally assigns the original ranking data to two classes. Then, the dimensionality of the data is reduced to 2-dimensional using PCA in order to display it on the plots of Fig. \ref{fig:dist}. The data is then labelled with two colors red and green indicating the relevance between samples. We can notice a high overlap between the two classes in the case of raw data (left plot of Fig. \ref{fig:dist}), while the data on the right is clearly better separated after the projection using the proposed MvMDAE. This shows the discrimination power of the proposed supervised embedding method. \par
\begin{figure}
[h!]
\centering
\includegraphics[width=.5\textwidth]{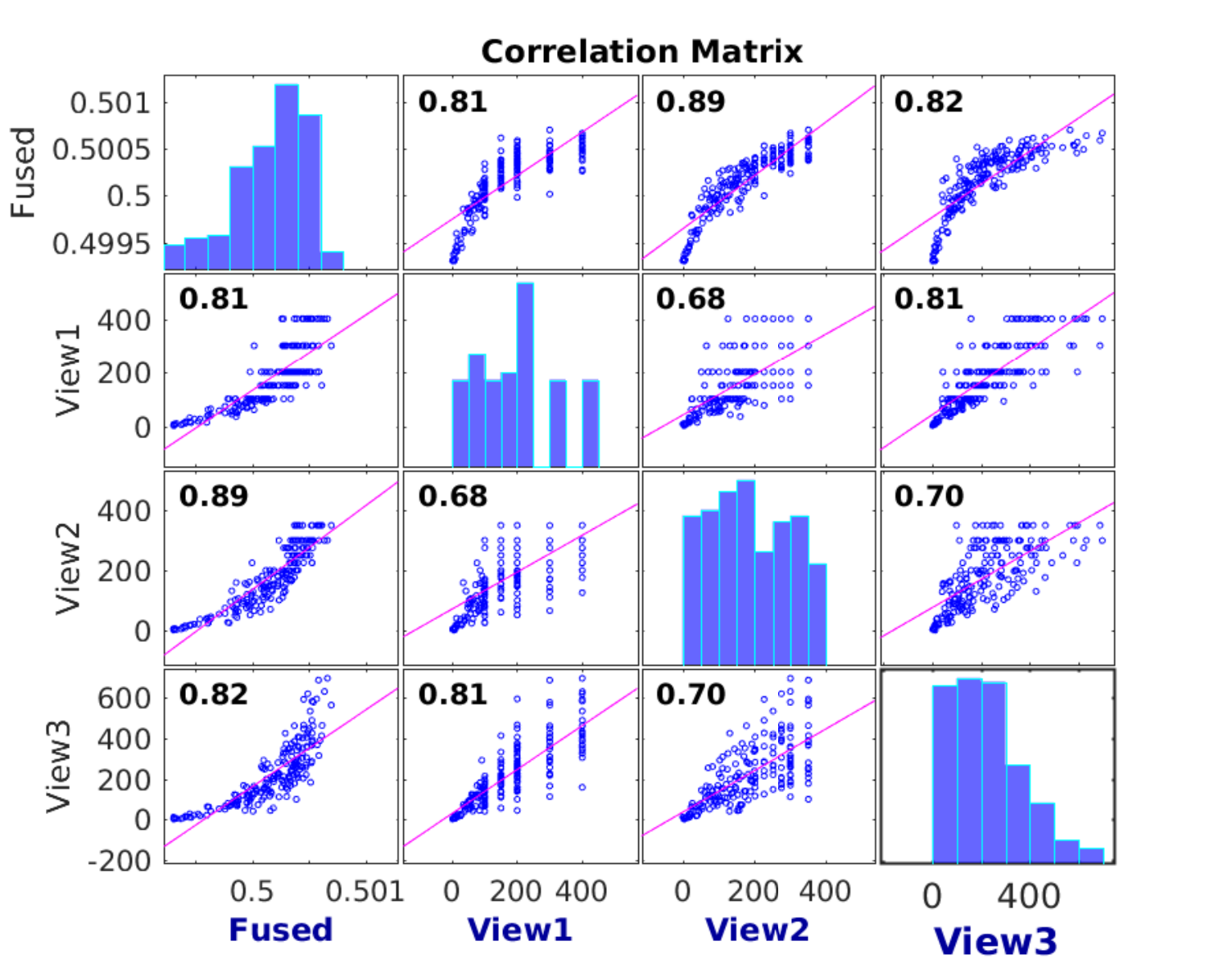}
\caption{Rank correlation matrix for views 1-3 and the fused view.}\label{fig:corr}
\end{figure}
Furthermore, a rank correlation matrix of plots is presented in Fig. \ref{fig:corr} with correlations among pairs of ranking lists from the views 1-3 and the predicted list denoted by 'Fused'. Histograms of the ranking data are shown along the matrix diagonal, while scatter plots of data pairs appear off diagonal. The slopes of the least-squares reference lines in the scatter plots are equal to the displayed correlation coefficients.
The fused ranking list is produced by the proposed DMvDR, and the results are also generated from the common universities in 2015. We first take a closer look at the correlations between the views 1-3. The correlation coefficients are generally low, with the highest ($0.81$) between view 1 and 3, while the others are around $0.70$. In contrast, the fused rank has a high correlation to each view. The scatter plots and the reference lines are well aligned, and the correlation coefficients are all above $0.80$, demonstrating that the proposed DMvDR effectively exploits the global agreement with all view.\par
Finally, the average prediction results over 3 different university datasets of the proposed and competing methods are reported in Table \ref{tab:univ}. Due to the misalignment of ranking data in 2015 across datasets, we make the ranking prediction based on each view input, which is further elaborated in the Section \ref{sec:dmvdr}. 
We observe that Ranking SVM \cite{Joachims2002} on the single feature or its concatenation performs poorly compared to the other methods. This shows that when the data is heterogeneous, simply combining the features cannot enhance joint ranking. Kendal's tau from the linear subspace learning methods are comparatively higher than their nonlinear counterparts. This is due to the fact that the nonlinear methods aim to maximize to the correlation in the embedding space, while the scoring function is not optimized for ranking. In contrast, DMvDR optimizes the entire ranking process, which is confirmed with the highest ranking and classification performance.
\begin{table}[h!]
\centering
\caption{Average Prediction Results (\%) on 3 University Ranking Datasets in 2015.}\label{tab:univ}
\begin{tabular}{l c c c c c c c c}
\\ \thickhline
Methods & Kendal's tau & Accuracy \\ \hline
Best Single View & 65.38  & - \\
Feature Concat & 35.10 &  -\\ 
LMvCCA \cite{Cao2017}& 86.04  & 94.49\\
LMvMDA \cite{Cao2017} & 87.00&  94.97\\
MvDA \cite{Kan2016} & 85.81 &    94.34\\
SmVR \cite{Usunier2011} & 80.75 & - \\
DMvCCA \cite{Cao2017} & 70.07 &  93.20\\
DMvMDA \cite{Cao2017} & 70.81 &  94.75\\
MvCCAE (\emph{ours}) & 75.94  &  94.01\\
MvMDAE (\emph{ours}) & 81.04 &  94.85\\
DMvDR (\emph{ours}) & \textbf{89.28} &  \textbf{95.30}\\ \thickhline 
\end{tabular}
\end{table}
\subsection{Multi-lingual Ranking}
\begin{figure}
[h!]
\centering
\begin{subfigure}[b]{\linewidth}
\includegraphics[width=.5\textwidth, height=4cm]{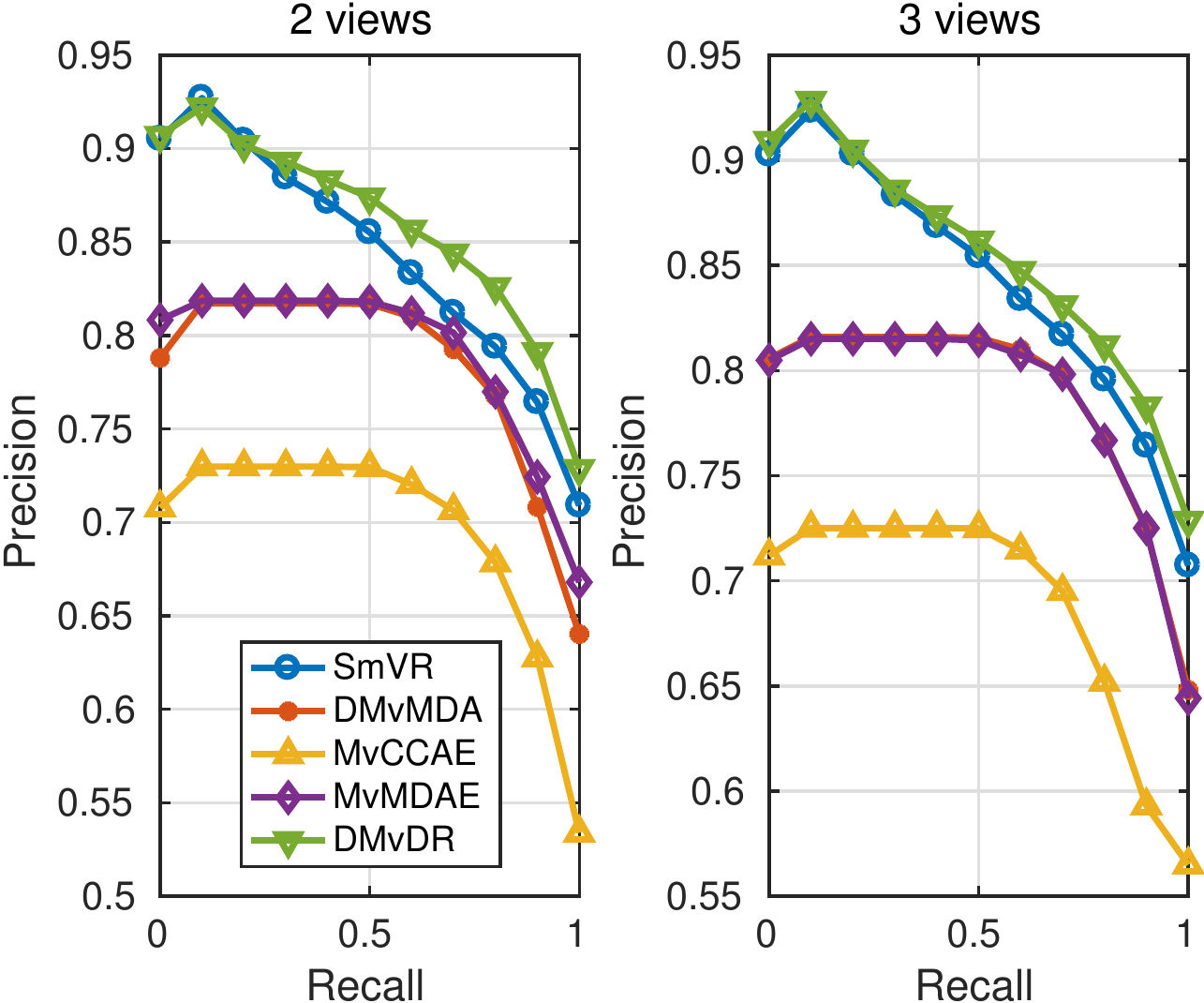}
\includegraphics[width=.5\textwidth, height=4cm]{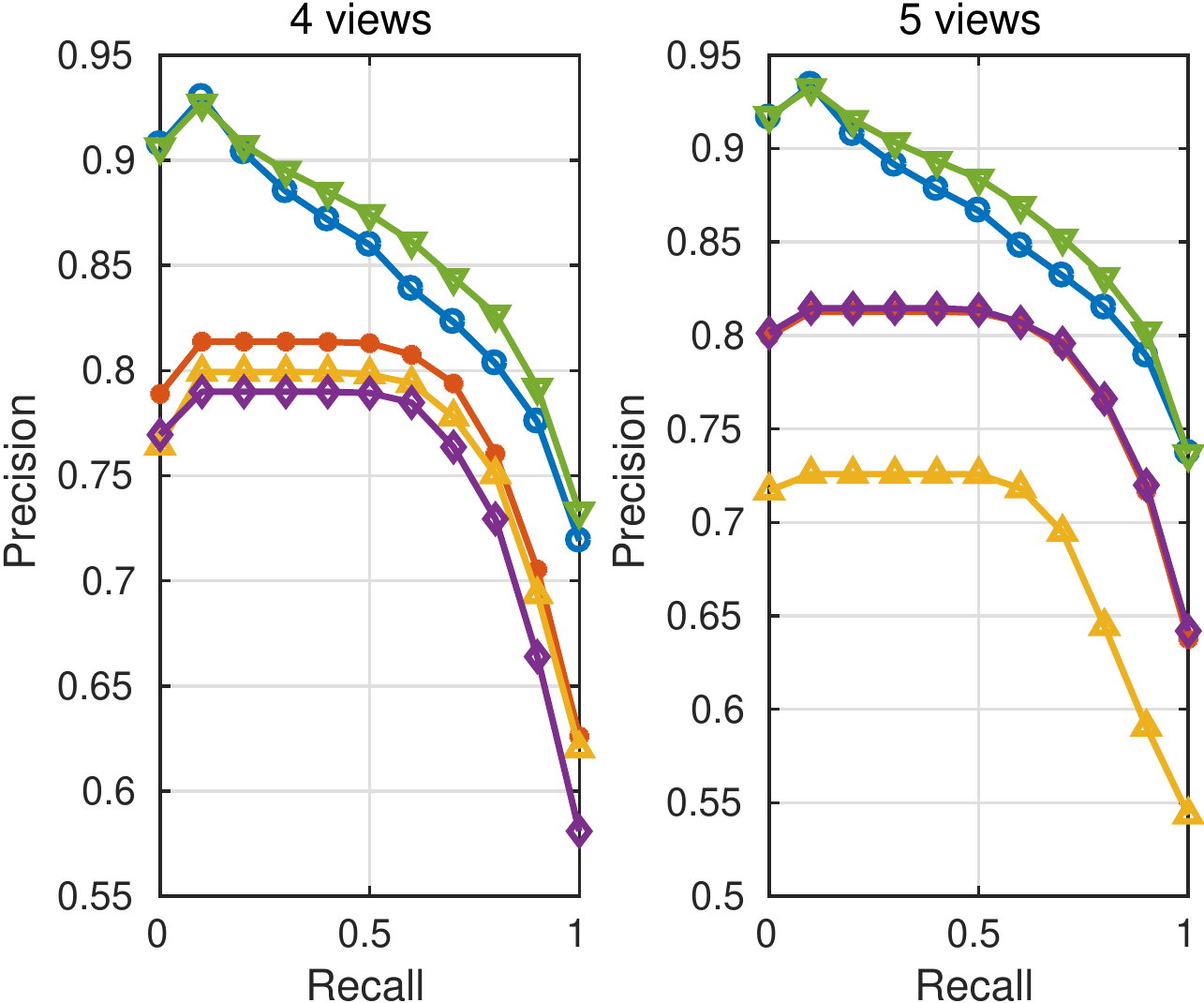}
\caption{PR curve on Reuters.}
\end{subfigure}\\[5mm]%
\begin{subfigure}[b]{\linewidth}
\includegraphics[width=.5\textwidth, height=4cm]{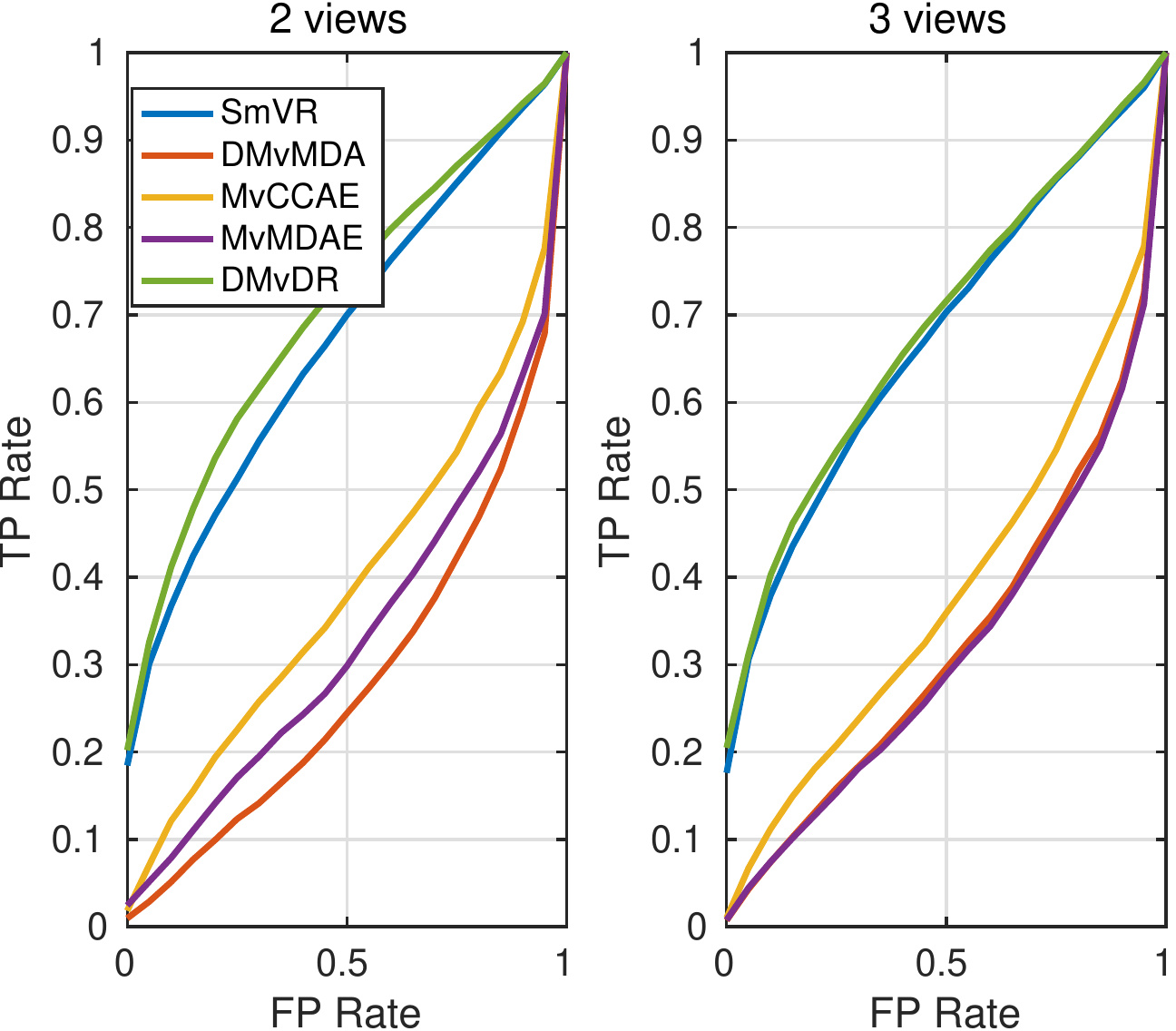}
\includegraphics[width=.5\textwidth, height=4cm]{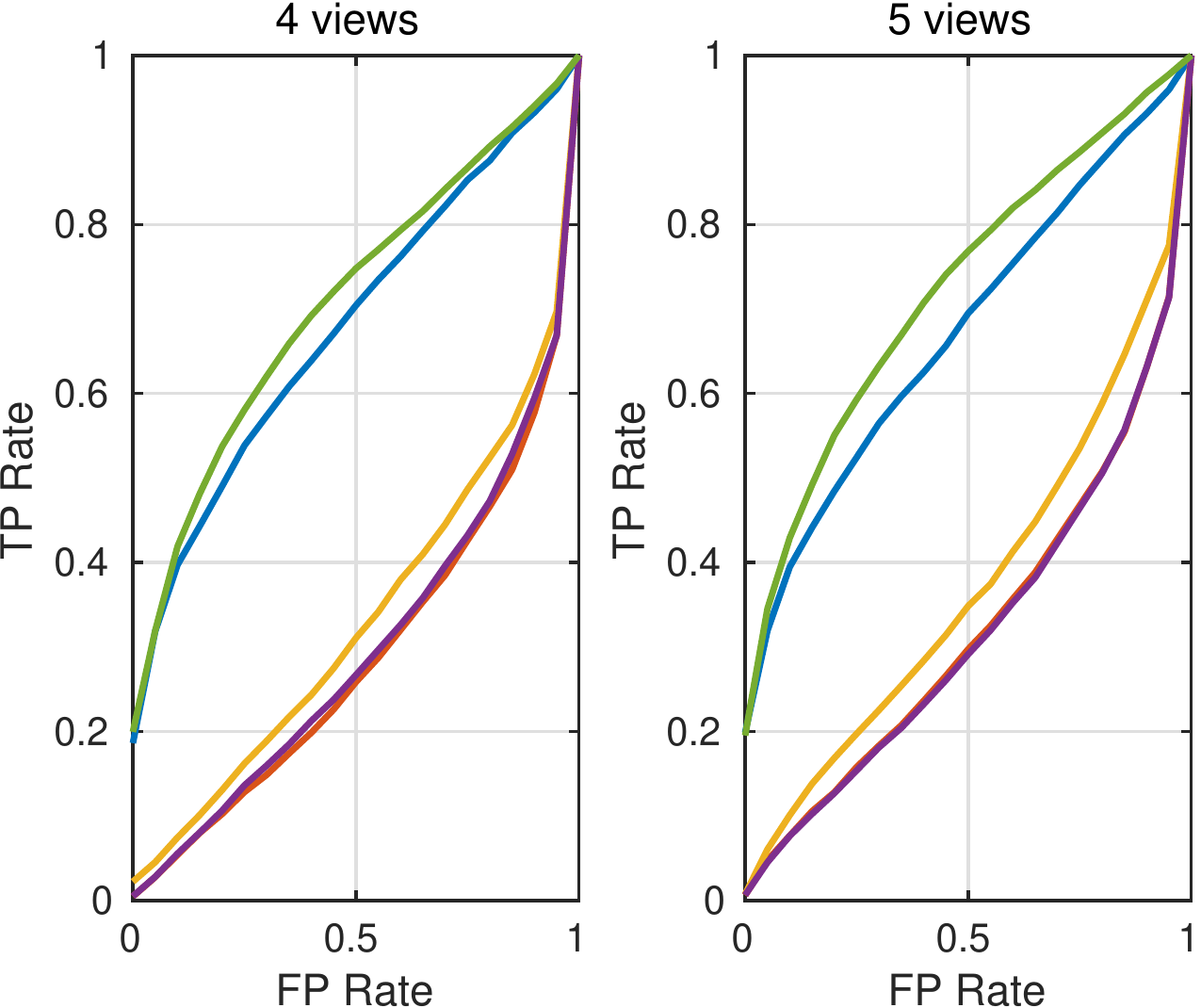}
\caption{ROC curve on Reuters.}
\end{subfigure}
\caption{The PR and ROC curves with 2-5 views applied to Reuters dataset.}\label{fig:reu}
\end{figure}
\begin{table*}[b!]
\centering
\caption{Quantitative Results (\%) on the Reuter Dataset.}\label{tab:reu}
\begin{tabular}{l c c c c c c c c}
\\ \thickhline
& \multicolumn{2}{ c }{2 views} & 
\multicolumn{2}{ c }{3 views} & 
\multicolumn{2}{ c }{4 views} & 
\multicolumn{2}{ c }{5 views} 
\\ \hline
Methods & MAP{@}100 & Accuracy &MAP{@}100 & Accuracy &MAP{@}100 & Accuracy &MAP{@} 100 & Accuracy \\ \hline 
Feature Concat & 58.87&  70.41&   56.97 &  70.10 &   57.59&   69.88&   58.46 &  69.97\\ 
LMvCCA \cite{Cao2017}& 59.10   &   70.20 &   62.40&  72.01 &  54.41 &  66.61&   60.41&   \textbf{72.62} \\
LMvMDA \cite{Cao2017} & 59.09 &  70.16  & 58.81&   71.94&   61.54&   72.45 &  59.28  & 72.07 \\
MvDA \cite{Kan2016} & 55.95 & 69.03 &  55.42 &  67.57 &  55.64 &  68.64 &  58.93 &  68.46 \\
SmVR \cite{Usunier2011} & 78.37 &71.44 &  78.24 &  71.15 &  78.66 &  71.37  & 79.36 &  71.64 \\
DMvCCA \cite{Cao2017} & 53.87 &  67.41 &  42.68&   62.02 &  54.51& 68.03 &  57.27 &  65.00 \\
DMvMDA \cite{Cao2017} & 60.08 &  71.40 &   63.12 &  70.93 &  61.55 &  72.12 &  62.52 &  70.78 \\
MvCCAE (\emph{ours}) & 48.75  &  66.43 &  49.10 &  62.90 &    60.70&    71.86&   48.80 &   63.05\\
MvMDAE (\emph{ours}) & 62.63 &   \textbf{74.20}&    63.02 &  71.04 &  60.74 &  72.60 &    62.74 &  71.20 \\
DMvDR (\emph{ours}) & \textbf{80.01} &    72.68 &  \textbf{79.34} &  \textbf{72.23}&   \textbf{80.32} &  \textbf{73.07} &  \textbf{81.64} &  72.39 \\ \thickhline 
\end{tabular}
\end{table*}
The Multi-lingual Ranking is performed on Reuters RCV1/RCV2 Multi-lingual, Multi-view Text Categorization Test collection \cite{Amini2009}. We use Reuters to indicate this dataset in later paragraphs. It is a large collection of documents with news ariticles written in five languages, and grouped into $6$ categories by topic. The bag of words (BOW) based on a TF-IDF weighting method \cite{Manning2008} is used to represent the documents. The vocabulary has a size of approximately $15000$ on average and is very sparse.\par
\par 
We consider the English documents and their translations to the other 4 languages in our experiment. Specifically, the 5 views are numbered as follows:
\begin{itemize}
\item View 1: original English documents; 
\item View 2: English documents translated to French; 
\item View 3: English documents translated to German;
\item View 4: English documents translated to Italian; 
\item View 5: English documents translated to Spanish.
\end{itemize}
Due to its high dimensionality, the BOW representation of each document is projected using a sparse SVD to a 50-dimensional compact feature vector. We randomly select 40 samples from each category in each view as training data. The training data composed of $28680$ samples is generated between pairs of English documents based on the pairwise transform in Section \ref{sec:pair}, and the translations to other languages are used for augmenting the views. We select another 360 samples from 6 categories and create a test dataset of $64620$ document pairs. If considering the ranking function linear as proved in \cite{Herbrich2000}, we make document pairs comparable and balance them by assigning some of the data to the other class with the opposite sign of the feature vectors, so that the number of samples is equally distributed in both classes. 
For MvCCAE and MvMDAE, the encoding networks $\F_v$ have three hidden layers, with $100$, $10$ and $10$ neurons each, while the decoding networks $\G_v$ have one hidden layer with $32$ neurons. Sigmoid activation function is used for all neurons in the networks. We used a minibatch size of $100$ samples, the learning rate is $10^{-3}$ and $\rho$ for $L_2$ penalty is set to $10^{-4}$. The encoding networks $\F_v$ in DMvDR have a topology of $50$ and $10$ hidden layer neurons, while the decoding networks $\G_v$ have a single hidden layer with $64$ neurons and one output neuron. $\Hb$ has the same topology as $\G_v$. We used a minibatch size of $200$ samples, the learning rate is $10^{-4}$ and $\rho$ for $L_2$ penalty is set to $10^{-2}$.\par
We first analyze the PR and ROC curves in Fig. \ref{fig:reu}. Since we have all translations of the English documents, each sample is well aligned in all views and, therefore we perform joint learning and prediction in all multi-lingual experiments. The experiments start with 2 views with English and its translation to French, and then the views are augmented with the documents of other languages. Subspace ranking methods are trained by embedding with increasing number of views, while SmVR as a co-training takes two views at a time, and the average performance of all pairs is reported.
The proposed methods with two competing ones are included in the plots in Fig. \ref{fig:reu}. The proposed DMvDR clearly performs the best across all views as can be seen in the PR and ROC plots in Fig. \ref{fig:reu}. SmVR is the second best with a lower precision and less area under curve compared to DMvDR. Among the remaining three methods, DMvMDA performs favorably in the PR curves but not as well in the ROC plots. The results are comparatively consistent across all views.\par

We can observe the quantitative MAP and accuracy results in Table \ref{tab:reu}.  It shows that the linear methods together with the feature concatenation have similar results which are generally inferior to the nonlinear methods in classification. Note also that nonlinear subspace learning methods cannot provide any superior MAP scores, which can be explained by the fact that the embedding is only intended to construct a discrimative feature space for classifying the pairs of data. 
We can also observe the MAP scores and accuracies are stable across views. This can be interpreted as a global ranking agreement can be reached to a certain level when all languages correspond to each other.
It is again confirmed that the end-to-end solution consistently provides the highest scores, while SvMR is a few percentages behind. When the features from different views follow a similar data distribution, the co-training method performs well and competes with the proposed DMvDR.
\subsection{Image Data Ranking}
Image data ranking is a problem to evaluate the relevance between two images represented by different types of features. We adopt the Animal With Attributes (AWA) dataset \cite{Lampert2014} for this problem due to its diversity of animal appearance and large number of classes. The dataset is composed of $50$ animal classes with a total of $30475$ images, and $85$ animal attributes. We follow the feature generation in \cite{Cao2017} to adopt 3 feature types forming the views:
\begin{itemize}
    \item Image Feature by VGG-16 pre-trained model: a 1000-dimensional feature vector is produced from each image by resizing them to $224\times 224$ and taken from the outputs of the $fc8$ layer with a 16-layer VGGNet \cite{Simonyan2015}.
    \item Class Label Encoding: a 100-dimensional Word2Vector is extracted from each class label. Then, we can map the visual feature of each image to the text feature space by using a ridge regressor with a similar setting as in \cite{Cao2017} to genenate another set of textual feature, with connection to the visual world. The text embedding space is constructed by training a skip-gram \cite{Mikolov2013} model on the entire English Wikipedia articles, including $2.9$ billion words.
\item Attibute Encoding: an 85-dimensional feature vetor can be produced with a similar idea as above. Since each class of animals contains some typical patterns of the attribute, a $50\times 85$ lookup table can be constructed to connect the classes and attributes \cite{Kemp2006, Osherson1991}. Then, we map each image feature to the attribute space to produce the mid-level feature. 
\end{itemize}
\par
\begin{figure}
[h!]
\centering
\includegraphics[width=.4\textwidth]{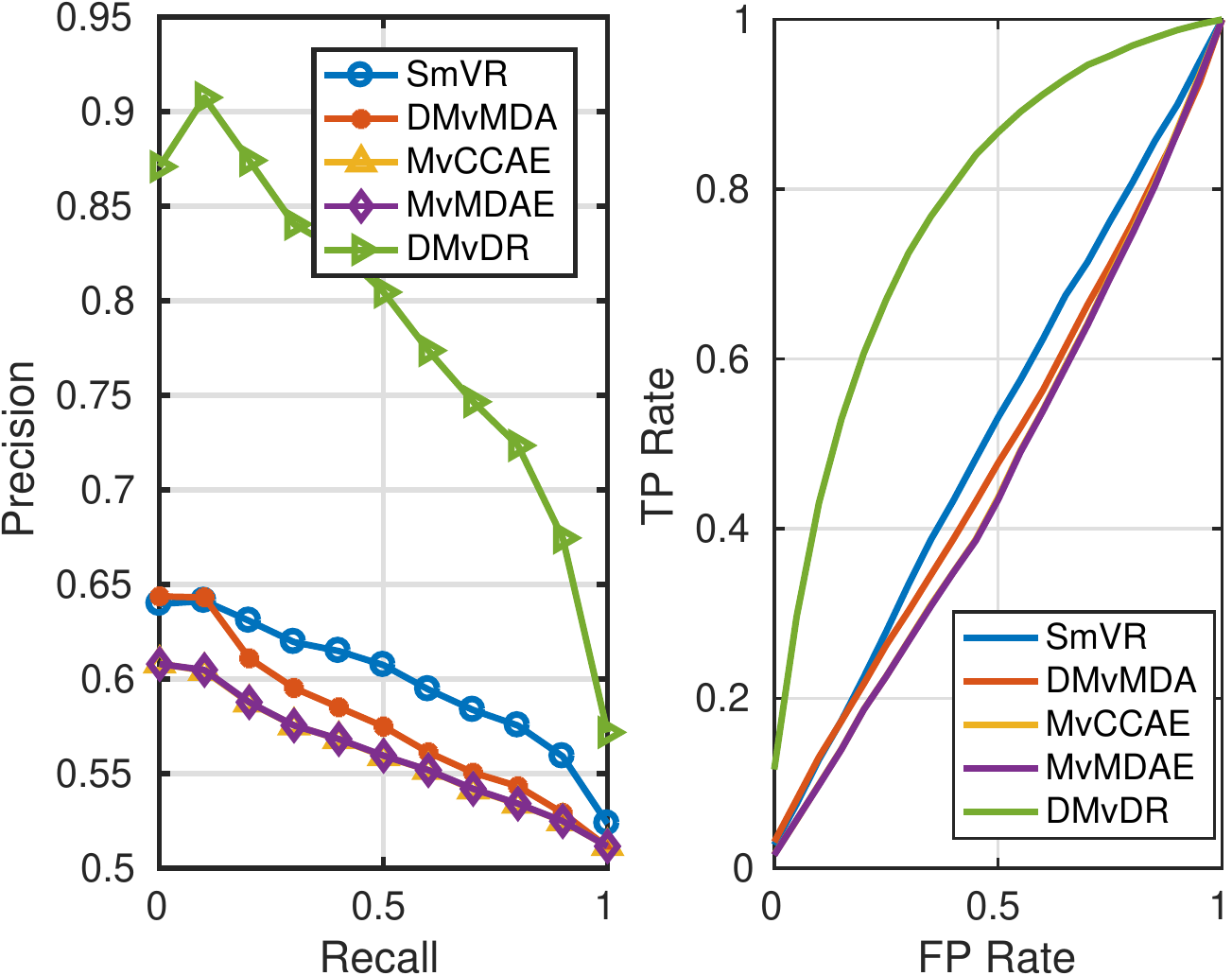}
\caption{PR and ROC curves on AWA.}\label{fig:awa}
\end{figure}
We generate the image ranking data as follows. From the 50 classes of animal images, we find $400$ pairs of images with $200$
in-class pairs and $200$ out-of-class image pairs from each class. We then end up with $20000$ training data pairs. Similarly, we will have $20000$ test data pairs.
We select $40$ images from each class used for training data and a separate set of $40$ samples as test data.
Another 10 images are used as queries: 5 of them are associated with the in-class images as positive sample pairs and 5 as negative sample pairs. For the negative sample pairs, we randomly select $40$ classes from $49$ remaining animal classes at a time, and one image per class is associated with each query image under study. 
We found a common topology for both MvCCAE and MvMDAE from the validation set. 
For MvCCAE and MvMDAE, the encoding networks $\F_v$ have two hidden layers, with $64$ and $10$ neurons each, while the decoding networks $\G_v$ have one hidden layer with $50$ neurons. Sigmoid activation function is used for all neurons in the networks. We used a minibatch size of $100$ samples, the learning rate is $10^{-4}$ and $\rho$ for $L_2$ penalty is set to $10^{-2}$. The encoding networks $\F_v$ in DMvDR have a topology of $100$, $100$ and $10$ hidden layer neurons, while the decoding networks $\G_v$ have a single hidden layer with $100$ neurons and one output neuron. $\Hb$ has the same topology as $\G_v$. We used a minibatch size of $200$ samples, the learning rate is $10^{-2}$ and $\rho$ for $L_2$ penalty is set to $10^{-8}$.\par 
\begin{table}[h!]
\centering
\caption{Quantitative Results (\%) on the AWA Dataset.}\label{tab:awa}
\begin{tabular}{l c c c c c c c c}
\\ \thickhline
Methods & MAP{@}100 & Accuracy \\ \hline
Feature Concat & 38.08 & 50.60 \\ 
LMvCCA \cite{Cao2017}&49.97 &  51.85 \\
LMvMDA \cite{Cao2017} & 49.70 &  52.35\\
MvDA \cite{Kan2016} & 49.20  & 52.82 \\
SmVR \cite{Usunier2011} & 52.12 & 50.33\\
DMvCCA \cite{Cao2017} & 51.38  & 50.83\\
DMvMDA \cite{Cao2017} & 51.52 &  51.38\\
MvCCAE (\emph{ours}) & 49.01 &   53.28\\
MvMDAE (\emph{ours}) & 48.99 &   53.30\\
DMvDR (\emph{ours}) & \textbf{76.83}  &   \textbf{71.48}\\ \thickhline 
\end{tabular}
\end{table}
We can observe the performance of the methods on the animal dataset graphically in Fig. \ref{fig:awa} and quantitatively in Table \ref{tab:awa}. DMvDR outperforms the other competing methods by a large margin as shown in the plots of Fig. \ref{fig:awa}. Due to the variety of data distribution from different feature types as view inputs, the co-training type of SmVR can no longer compete with the end-to-end solution. From Table \ref{tab:awa}, one can observe that the performance of the feature concatenation suffers from the same problem. On the other hand, our proposed subspace ranking methods produces satisfactory classification rates while the precisions remain somewhat low. This implies again the scoring function is critical to be trained together with the feature mappings. The other linear and nonlinear subspace ranking methods have comparatively similar performance at a lower position.\par
\begin{figure}
[h!]
\includegraphics[width=.45\textwidth]{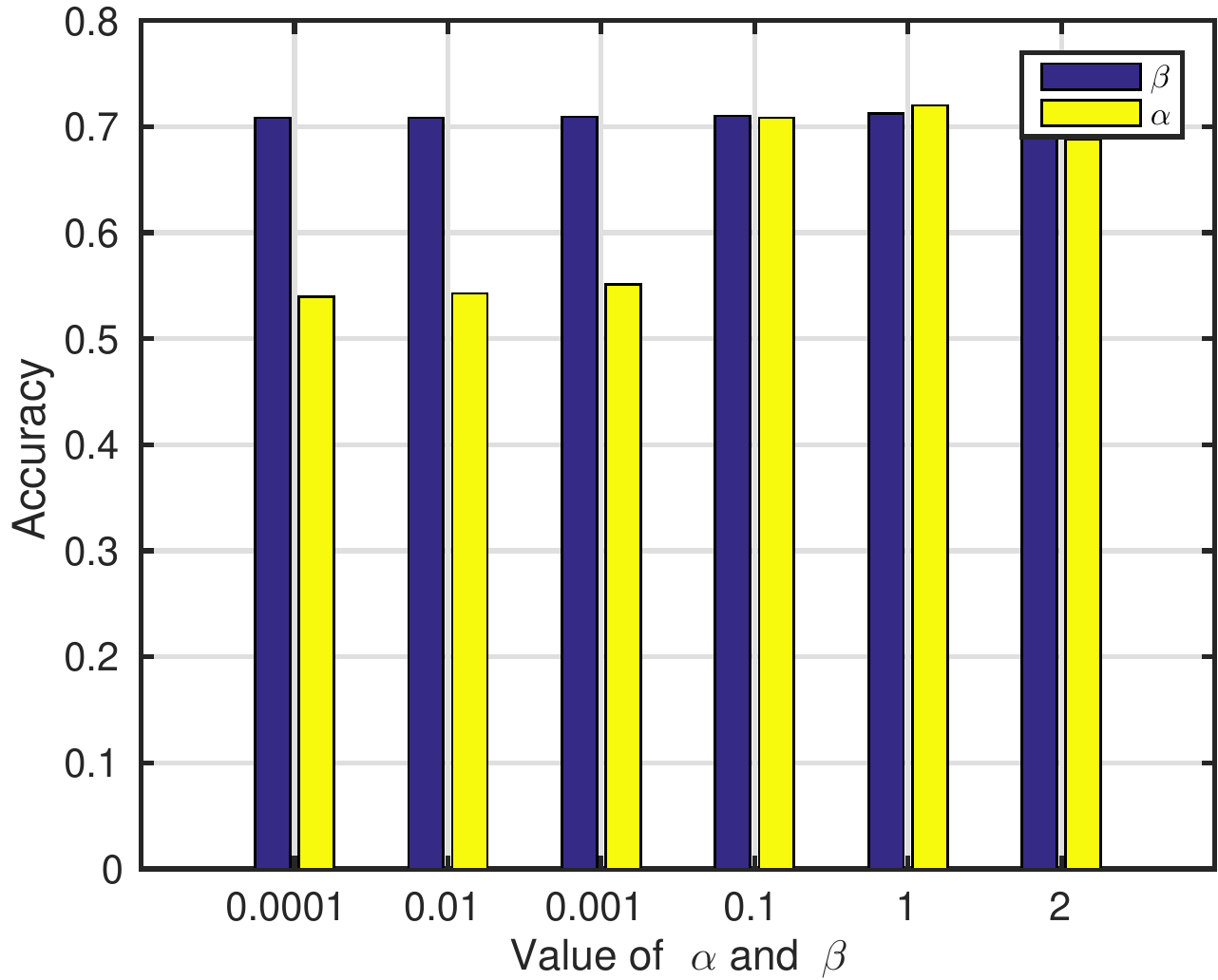}
\caption{Performance of DMvDR on different values of $\alpha$ and $\beta$.}\label{fig:param}
\end{figure}
We also study the effects of selecting different values of $\alpha$ and $\beta$ during training on the classification performance of DMvDR. The results are shown in Fig. \ref{fig:param}. One parameter varies across the neural network models while the other one is fixed based on a grid search. While the performance is mostly consistent with the value changes of $\alpha$ and $\beta$, it drops when $\alpha$ is below $0.001$. It shows the importance of jointly optimizing the view-specific ranking. \par
\section{Conclusion}\label{sec:cl}
Learning to rank has been a popular research topic with numerous applications, while multi-view ranking remains a relatively new research topic. In this paper, we aimed to associate the multi-view subspace learning methods with the ranking problem and proposed three methods in this direction. 
MvCCAE is an unsupervised multi-view embedding method, while MvMDAE is its supervised counterpart. Both of them incorporate multiple objectives, with a correlation maximization on one hand, and reconstruction error minimization on the other hand, and have been extended in the multi-view subspace learning to rank scheme. Finally, DMvDR is proposed to exploit the global agreement while minimizing the individual ranking losses in a single optimization process. The experimental results validate the superior performance of DMvDR compared to the other subspace and co-training methods on multi-view datasets with both homogeneous and heterogeneous data representations. \par
In the future, we will explore the scenario when there exists missing data, which is beyond the scope of the current proposed subspace ranking methods during training. Multiple networks can also be combined by concatenating their outputs, and further optimized in a single sub-network. This solution may also be applicable for homogeneous representations.
\bibliographystyle{IEEEtran}
\bibliography{ResearchBib.bib}
\end{document}